%% file: main.tex
\pdfoutput=1

\documentclass[11pt]{article}

\usepackage[dvipsnames]{xcolor}

\usepackage[final]{acl}

\usepackage{times}
\usepackage{latexsym}

\usepackage{}
\usepackage[utf8]{inputenc} 
\usepackage[T1]{fontenc}    
\usepackage{hyperref}       
\usepackage{url}            
\usepackage{booktabs}       
\usepackage{amsfonts}       
\usepackage{nicefrac}       
\usepackage{microtype}      
\usepackage{xcolor}         
\usepackage{multirow}
\usepackage{algorithm}
\usepackage{algpseudocode}
\usepackage{amsmath}
\usepackage{amsthm}
\usepackage{booktabs}
\usepackage{enumitem}
\usepackage{hhline}
\usepackage{multirow}
\usepackage{pdfpages}
\usepackage{pifont}
\usepackage{setspace}
\usepackage{soul}
\usepackage{tabularx}
\usepackage{graphicx}
\usepackage{url}
\usepackage{todonotes}
\title{Fine-grained Text Style Transfer with Diffusion-Based Language Models }

\begin{document}

\author{Yiwei Lyu \quad Tiange Luo \quad Jiacheng Shi \quad Todd C. Hollon \quad Honglak Lee\\
\\
University of Michigan\\
\{yiweilyu, tiangel, jiachs\}@umich.edu\\
}


\maketitle

\begin{abstract}
Diffusion probabilistic models have shown great success in generating high-quality images controllably, and researchers have tried to utilize this controllability into text generation domain. Previous works on diffusion-based language models have shown that they can be trained without external knowledge (such as pre-trained weights) and still achieve stable performance and controllability. In this paper, we trained a diffusion-based model on StylePTB dataset, the standard benchmark for fine-grained text style transfers. The tasks in StylePTB requires much more refined control over the output text compared to tasks evaluated in previous works, and our model was able to achieve state-of-the-art performance on StylePTB on both individual and compositional transfers. Moreover, our model, trained on limited data from StylePTB without external knowledge, outperforms previous works that utilized pretrained weights, embeddings, and external grammar parsers, and this may indicate that diffusion-based language models have great potential under low-resource settings. Our code is available at \url{https://github.com/lvyiwei1/DiffuSeq\_StylePTB}

\end{abstract}

\section{Introduction}
%

\input{tables/example}

Diffusion probabilistic models~\cite{ho2020denoising} have became the state-of-the-art technique in visual generative tasks. By starting from random gaussian noise and gradual denoising, they are able to generate images that look realistic in details. Moreover, conditional diffusion models such as stable diffusion~\cite{rombach2022high} are able to achieve detailed control over the generated output by conditioning on text, layouts, etc. The generated images are faithful to the text description or layouts, often to the finest details. 

Analogically, researchers have tried to utilize the controllability of diffusion models to achieve more controllable language generation. For example,  DiffuSeq~\cite{gong2022diffuseq} applies diffusion models to sequence-sequence text generation tasks such as paraphrasing, question generation and text simplification; Diffusion-LM~\cite{li2022diffusion} combined diffusion models with language models to control language generation by specifying generation length, syntax tree, semantic context, etc. 
What made these diffusion-based language models impressive is that they are trained from scratch with zero external knowledge (i.e. no pre-trained word embeddings or model weights, no external grammar parsers, etc) and on very few data (on the order of $10^5$ tokens) compared to any large language models (for example, GPT-3's~\cite{brown2020language} training data is on the order $10^{11}$ tokens), so they have to learn representations at all levels (word embeddings, sentence structures, etc) from scratch with very limited data.

However, while the earlier tasks assessed on Diffusion-LM and DiffuSeq require a degree of control over the generated output, they are incapable of modifying the existing text to exhibit specific stylistic characteristics. In this paper, we would like to further examine the capabilities of diffusion-based language models on \textbf{fine-grained text style transfer}, an important task that requires more fine-grained control than the tasks from previous works on diffusion-based language modeling because it only allows changing the specified fine-grained stylistic properties of the input while leaving the rest unchanged. For example, "verb emphasis" is a fine-grained style transfer that requires the model to rewrite the sentence emphasizing a certain verb, without changing any other information that the original sentence conveys. In comparison, previous evaluation tasks such as controlling sequence length, semantic context, etc essentially control one aspect at a time and require no control over any other properties of generated text. 


We use 13 non-lexical transfers from StylePTB~\cite{lyu2021styleptb} dataset, where there are at most a few thousand sentence pairs available for each transfer, as shown in Table~\ref{tab:example}. Since identifying the grammatical structure of the sentence can be very helpful for most of these transfers (such as active-to-passive), some previous methods (such as Neural QCFG~\cite{kim2021sequence}) utilizes external grammar parsers to gain such information. We trained a diffusion-based model on StylePTB data without any pre-trained weights or external grammar parsers. Therefore, our model has to start from zero grammar/linguistic knowledge and learn all of them from very limited training data (StylePTB only has 7719 sentences from Penn Tree Bank~\cite{marcus-etal-1993-building} plus their transferred outputs). Even under these hard conditions, our model still managed to outperform previous works that do utilize external weights or grammar parsers. Moreover, we also evaluate the capabilities of diffusion-based language models on performing multiple transfers using one single model and composing multiple learned transfers on a single sentence. We list our contributions as follows:
\vspace{-0.5mm}
\begin{itemize}[itemsep=0em]
    \item We trained a diffusion-based language model (adapted from DiffuSeq~\cite{gong2022diffuseq}) that can perform fine-grained text style transfer from scratch with very limited training data and no external weights or tools. The model also supports multitasking and composing multiple fine-grained transfers. 
    
    \item Our model achieves state-of-the-art 
    performance on fine-grained text style transfers in StylePTB. Our multitask model (i.e. one single model that can perform all 13 transfers) achieves \textbf{best performance} compared to previous works on the same tasks on \textbf{88 out of 91} metrics (7 metrics per transfer), and gets very close to human performance on tasks with easy and medium difficulties. We also evaluated our model on composition of multiple fine-grained transfers, and we achieved best performance on these tasks as well.
    \item Thr/ough the evaluations, we demonstrated the extraordinary capabilities of diffusion-based language models in asserting extremely fine-grained control over generated text, and that this type of language model have great potential in controllable natural language generation under low-resource settings as it is able to achieve state-of-the-art performance with limited training data and no external knowledge.
\end{itemize}

\section{Backgrounds}

\subsection{Fine-grained Text Style Transfer and StylePTB}

An import challenge for AI is to convey intentions using different stylistic attributes, and automated text style transfer is an essential step towards that. Text style transfer aims to controllably convert source text with targeted stylistic properties, with important applications in human-AI interactions including dialog systems~\citep{celikyilmaz2018deep} and intelligent agents~\citep{kim2013social,liang2020emergent,Pittermann2010} that can communicate with specific text styles for different situations, target audiences, and environments~\citep{lample2018multipleattribute,li2018delete}. 

There has been extensive research on high-level style transfers such as sentiment transfers~\cite{shen2017style} and formality transfers~\cite{rao2018dear}. However, high-level style transfers lack the ability to fully control the style of the output. For example, there are many ways to convert a positive comment about a restaurant into a negative one, and high-level text style transfers do not allow control over which of the possible outputs (that may have different styles in non-sentiment aspects) can be generated. Fine-grained text style transfer is important because they allow fine-grained control over the generated output. \cite{lyu2021styleptb} defined a set of fine-grained text style transfer along four lingustic axis:  

\begin{itemize}[itemsep=0em]
    \item \textbf{Lexical Transfers: } Word changes\vspace{-1mm}
    \item \textbf{Syntax Transfers: } Grammar and sentence structure changes\vspace{-1mm}
    \item \textbf{Semantic Transfers: } Meaning changes \vspace{-1mm}
    \item \textbf{Thematic Transfers: } Situational changes or word emphasis
\end{itemize}


Along these 4 axes, it defined 21 individual fine-grained transfers, 13 of which are non-lexical. Examples of the non-lexical transfers are shown in Table~\ref{tab:example}.
 Compared to other forms of controllable text generation, fine-grained text style transfer has the advantage of being able to assert control over text generated by uncontrollable models. For example, we can use fine-grained text style transfers to add specific stylistic properties to free-form text generated by large language models while keeping the content of the generated text unchanged. Fine-grained text style transfers can be composed to achieve higher-level style transfers, and they even have the potential to mitigate social bias in large text generation models~\cite{lyu2021styleptb}.
Therefore, it is important to develop techniques to achieve automated fine-grained text style transfer. Existing works are still quite far from perfect on a lot of the fine-grained style transfers compared to human performance~\cite{lyu2021styleptb,kim2021sequence}, and composing multiple fine-grained style transfers remains challenging.

\vspace{-1mm}
\subsection{Diffusion Probabilistic Models}
\vspace{-1mm}
Recently, diffusion models~\cite{ho2020denoising} is widely used to generate high quality and diverse images. Its methodology consists of two phases: the first phase is the forward diffusion phase, which adds Gaussian noise to the input image $x_0$ as the time stamp increases, and after enough steps the image is reduced to pure Gaussian noise $x_t$. The second phase is the recovery phase, in which a model is trained to gradually remove noise from $x_t$ until it recovers the original image $x_0$. During inference, we start from a randomly sampled gaussian noise $x_t$ and use the denoising model to gradually infer an image $x_0$.

Diffusion-based language generation models follows a similar approach where we perform the diffusion and denoising process in the token embedding space. We will explain the model we use, which is built upon DiffuSeq~\cite{gong2022diffuseq}, in details in the next section.



\vspace{-1mm}
\section{Methodology}
\vspace{-1mm}
\begin{figure*}[th]
    \centering
    \includegraphics[width = 1.0\textwidth]{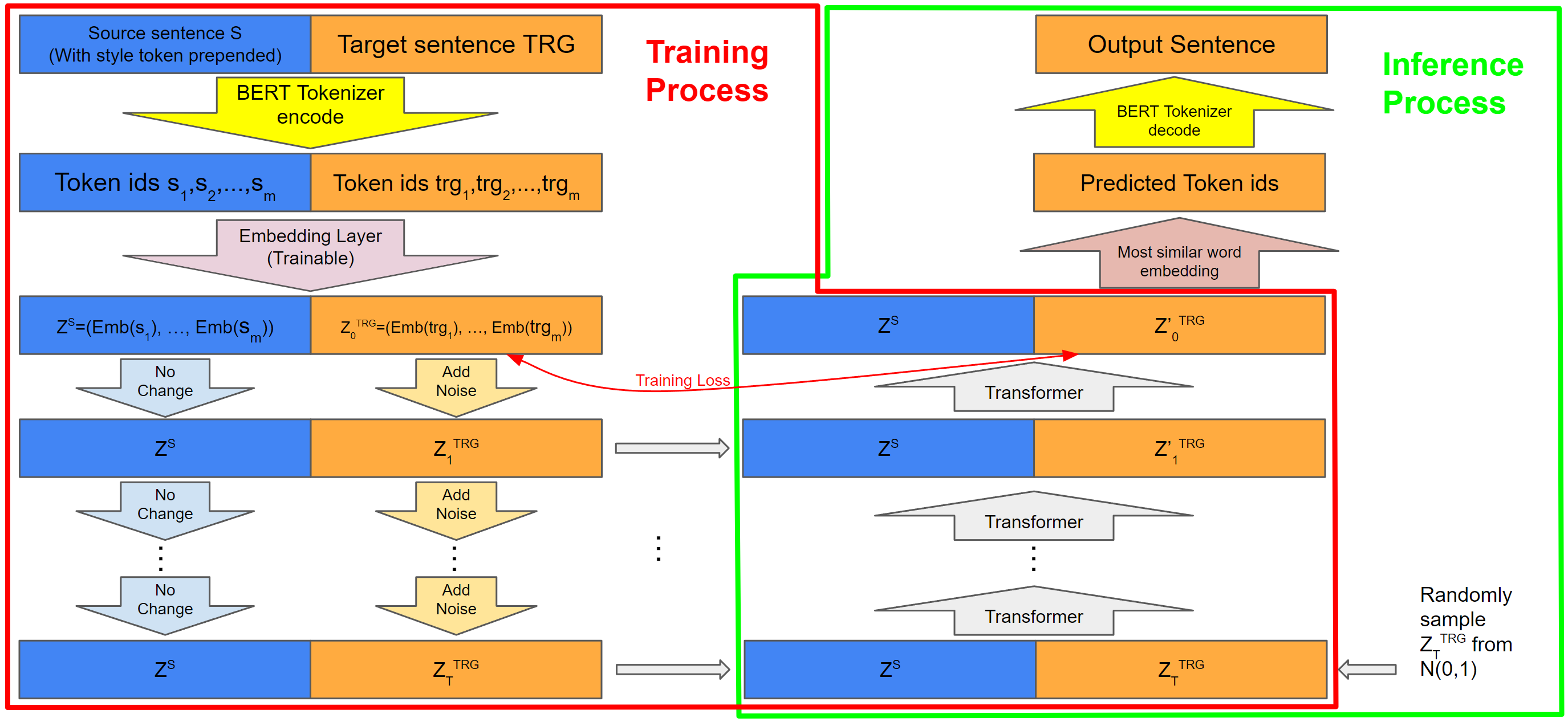}
    \caption{An illustration of the training and inference process of our diffusion-based language model. The diffusion process is performed over the sequence of token embeddings of the target sentence $Z_0^{TRG}$, and the source sentence's token embeddings ($Z^S$) are concatenated before $Z^{TRG}$. During the backward diffusion process, the combined sequence is fed into the transformer model to gradually recover/generate $Z_0^{TRG}$.\vspace{-5mm}}
    \label{fig:overview}
\end{figure*}

We adapt DiffuSeq~\cite{gong2022diffuseq} to be able to perform fine-grained text style transfer given a source sentence and specified transfer operation(s), as illustrated in Figure~\ref{fig:overview}. We model the transfer as a conditional generation process, where the condition includes the source sentence and the specified transfer operation(s). We first define a set of special style tokens, one for each possible individual fine-grained transfer. If we wish to perform one or more transfer on the source sentence, we will prepend the corresponding special token(s) to the beginning of the source sentence to form the condition $S$. 

We use BERT tokenizer to tokenize the input into discrete token ids, and adopt a token embedding layer to encode both the source (including prepended style tokens) and the ground truth target sentence (during training) to obtain the embedded source $Z^S$ and target $Z^{TRG}_0$. For the diffusion process, we use a transformer model to recover the target embedding. Both the diffusion transformer and the token embeddings are initialized randomly and jointly optimized. In other words, our model does not rely on any prior knowledge about our task or the English Language in general.

We use the simplified diffusion objective during training: for each input $(S,TRG)$ where $S$ is the source sentence (with style tokens) and $TRG$ is the ground truth target sentence, we randomly sample a step number $t$ from $1,2,... T$, where $T$ is the maximum number of steps, and add $t$ steps of random Gaussian noise to $Z^{TRG}_0$ following a linear diffusion schedule to obtain $Z^{TRG}_t$. We then concatenate $Z^S$ and $Z^{TRG}_t$ and input the concatenated sequence into our diffusion transformer, where we only take the output embeddings at the locations corresponding to $Z^{TRG}_t$ as $Z'^{TRG}_0$. Our training objective is simply going to be the MSE Loss between $Z^{TRG}_0$ and $Z'^{TRG}_0$.

During inference, we randomly initialize $Z'^{TRG}_T \sim N(0,1)$, and encode the condition (source sentence and style tokens) into $Z^S$. Then we concatenate them and use our transformer to predict a temporary $Z'^{TRG}_{0_{temp}}$, add $T-1$ steps of noise back to the temporary $Z'^{TRG}_{0_{temp}}$ to obtain $Z'^{TRG}_{T-1}$. We repeat this process until we get $Z'^{TRG}_0$. For each embedding in $Z'^{TRG}_0$, we find the closest embedding in our token embedding layer by cosine distance, and decode the embedding to that token. Then we combine the tokens to form the output sentence in natural language.

\section{Experiments}

\subsection{Dataset}

StylePTB~\cite{lyu2021styleptb} contains paired sentences before/after each transfer for 21 fine-grained transfers, as well as paired data for compositions of multiple fine-grained transfers. For single transfers, we will focus on the 13 non-lexical fine-grained style transfers following~\cite{lyu2021styleptb}. The number of sentence pairs available from StylePTB for each transfer and examples of sentences before/after each transfer are shown in Table~\ref{tab:example}. For compositional transfers, we will use the Tense + Voice and Tense + PP Removal transfers from the compositional part of StylePTB dataset (same as the ones used for evaluation in~\cite{lyu2021styleptb}). Each compositional dataset contains all combinations of valid transfers (for example, Tense + Voice dataset contains all valid combinations of 0/1/2 transfers regarding tense and voice, such as To-Future + Active-To-Passive or To-Past + No-Voice-Change). 

StylePTB was built with only 7719 different sentences from Penn Tree Bank~\cite{marcus-etal-1993-building} plus their stylistic variations, so both the amount and the diversity of training data are very limited, thus making this task even more challenging for DiffuSeq since it does not have access to external knowledge or pre-trained weights and have to extract all linguistic knowledge from limited data.

For fair comparison, we preprocess the data following the same criterion as~\cite{lyu2021styleptb}: we replace numbers with NUM token, and we replace each word that occurs less than 3 times in the training set with UNK token. We also split the data into train/valid/test splits with proportions of 0.9/0.05/0.05 using the same splits as all previous works.
\subsection{Evaluation Metrics}
We use the same evaluation methods as ~\cite{lyu2021styleptb} and report 7 metrics from nlg-eval package~\cite{sharma2017nlgeval} (BLEU 1-4, METEOR, ROUGE-L, CiDER) between the generated transferred sentence and the ground truth target sentence from the dataset.
\input{tables/res_short.tex}
\input{tables/full_hard.tex}
\input{tables/comp_full.tex}
\subsection{Single style transfer experiment}
\subsubsection {Baselines}

We report performance of the following baselines for single style transfer:

\vspace{-1.2mm}
\begin{enumerate}[itemsep=0em]
    \item \textbf{GPT-2}: Directly finetuning GPT-2 medium model~\cite{radford2019language} with paired data. Performance reported from ~\cite{lyu2021styleptb}.
    
\vspace{-1.2mm}
    \item \textbf{Seq2Seq}: GRU sequence-to-sequence language model~\cite{sutskever2014sequence} with attention. Performance reported from ~\cite{lyu2021styleptb}.
    
\vspace{-1.2mm}
    \item \textbf{RetrieveEdit~\cite{hashimoto2018retrieve}}: For an input data $x$, a retriever model will go through the training set to find a similar sentence pair $(x',y')$ and a trained editor edits $y'$ into desired output $y$. Performance reported from ~\cite{lyu2021styleptb}.
    
\vspace{-1.2mm}
    \item \textbf{Steering Vector~\cite{subramani2022extracting}}: extract steering vectors directly from pretrained LMs to guide generation
    
\vspace{-1.2mm}
    \item \textbf{TAILOR~\cite{ross2021tailor}}: output sentences conditioned on control codes by a pretrained seq2seq model
    \item \textbf{Neural QCFG~\cite{kim2021sequence}}: It presents a sequence-to-sequence text learning by explicitly modeling the alignment between target trees with the source.
    
\vspace{-1.4mm}
    \item \textbf{Neural QCFG + copy~\cite{kim2021sequence}}: Neural QCFG with an option to copy certain tokens from source sentence
\end{enumerate}


Among these baselines, \textbf{GPT-2}, \textbf{Steering Vector} and \textbf{TAILOR} uses pre-trained language models, \textbf{Neural QCFG} and \textbf{Neural QCFG + copy} requires external grammar parsers, and \textbf{RetrieveEdit} uses GLOVE word embeddings.

We also included \textbf{Human} performance on these tasks (reported in~\cite{lyu2021styleptb} by asking human annotators to manually perform the style transfer tasks) for comparison. 

\vspace{-1mm}
\subsubsection{Results and Analysis}

For single style transfers, we tried two different diffusion-based approaches: (1) we train a separate diffusion model for each individual style transfer, and (2) we train one diffusion model for all 13 transfers evaluated. For approach (2), we add a style token at the beginning of the input sentence to indicate which of the 13 transfers needs to be performed. We call approach (2) DiffuSeq Multitask.

The original StylePTB paper~\cite{lyu2021styleptb} puts the non-lexical transfers into 3 difficulty categories (easy, medium, hard) by average hamming distance between input and output of the transfer. We report the results of our experiment using the same categorization, where we show results on easy and medium transfers in Table~\ref{tab:bas_short} and hard transfers in Table~\ref{tab:full_shard}.

Surprisingly, DiffuSeq Multitask outperforms DiffuSeq on all transfers, even though DiffuSeq Multitask has to handle 13 different transfers in one model while each DiffuSeq model only needs to handle 1 transfer. This is possibly due to the additional training data from all the tasks that the multitask model learns better representations for words and sentences and gains more accurate knowledge of grammatical patterns of English, which is shared across all tasks.

Moreover, DiffuSeq Multitask significantly outperforms all baselines in all easy and medium transfers, and also achieves state-of-the-art on most metrics on hard transfers, only falling slightly behind Neural QCFG + copy in some metrics. This is really impressive considering that our approach leverages no external knowledge while all baselines except Seq2Seq utilizes either pretrained language models, pretrained word embeddings, or external grammar tree parser. Neural-QCFG-based methods are especially dependent on external linguistics knowledge and existing grammar parsers. DiffuSeq Multitask's performance is also on par with human performance on easy and medium transfers, indicating that DiffuSeq Multitask is close to fully solving the easy and medium difficulty transfers.
\subsection{Compositional style transfer experiment}
\subsubsection{Baselines}
We will report performance of the following baselines for compositional fine-grained style transfers:
\vspace{-3mm}
\begin{enumerate}[itemsep=0em]
    \item \textbf{SeqGPT}: Sequentially applying fine-tuned GPT-2 for each single style transfer. Performance reported from ~\cite{lyu2021styleptb}.
    \vspace{-1mm}
    \item \textbf{CS-GPT}: A modified GPT-2 model that takes in style tokens as indication of which style transfers to apply. Performance reported from ~\cite{lyu2021styleptb}.
    
\end{enumerate}
\subsubsection{Results and Analysis}

For compositions of multiple fine-grained style transfers, we train one single DiffuSeq model to handle all compositions and use style tokens to indicate which transfers to compose for the input sentence, similar to CS-GPT~\cite{lyu2021styleptb}. The results are shown in Table~\ref{tab:comp_full}. DiffuSeq significantly outperforms baselines in all tasks and all metrics. Therefore, not only does our diffusion model work well for single fine-grained style transfers, it also works well for compositions of multiple fine-grained style transfers.
\section{Related Works}
\subsection{Automated Text Style Transfer}

The goal of the text style transfer (TST) task is to change the style of the sentence while retaining its style-independent content. Previous works in TST includes the following approaches: statistical NLP methods~\cite{hovy1987generating,xu2012paraphrasing}, neural generative models~\cite{prabhumoye2018style,lample2018multipleattribute,he2020probabilistic}, Retrieve-and-Edit approaches~\citep{li2018delete,hashimoto2018retrieve,guu2018generating,sudhakar2019transforming,madaan2020politeness}, and Transformer-based approach~\cite{lyu2021styleptb}. Some of these methods can already achieve high performance on certain high-level transfers (such as sentiment transfers~\cite{shen2017style} and formality transfers~\cite{rao2018dear}), but fine-grained text style tranfer remains challenging for the above approaches~\cite{lyu2021styleptb}. In this paper, we explored a new approach for fine-grained TST utilizing Diffusion Models.

\subsection{Natural language processing with diffusion model}
There have been two approaches for leveraging diffusion models into text data: the first approach takes advantage of the diffusion model in the continuous domain, like Diffusion-LM~\cite{li2022diffusion}, and DiffuSeq~\cite{gong2022diffuseq}, where we start from a gaussian noise vector, and gradually denoise this noise vector to the desired sentence; the second approach applies diffusion model into discrete state space, like Multinomial Diffusion~\cite{hoogeboom2021argmax}, DDPMs~\cite{austin2021structured}, and DiffusionBERT~\cite{austin2021structured}. In this paper, we chose to build upon the first type of model, because they are closer to the original diffusion models for images (where diffusion happens in continuous space) and they have shown successes on tasks that requires control over generations.
\section{Limitations and Future works}
One significant limitation of our work is that we only explored the capabilities of diffusion-based language models under a challenging circumstance where it is not allowed to use pre-trained weights or grammar parsers, which means we did not utilize this kind of model to its full potential, so a future research direction could be exploring possible ways to further improve the model's performance by leveraging pretrained weights or word embeddings, and train with enough data to find the full potential of these models. 

Another limitation of our work is that we only explored one typical diffusion-based language model, so our conclusions may not generalize to special types of diffusion-based language models (such as ones that uses discrete state space). We also conducted all experiments using the exact same model architecture design. In the future, we plan to experiment with different architectures for the diffusion model, such as more sophisticated conditioning methods (currently we just concatenate the source to the target, but we would like to try other ways of conditioning on the source, such as cross attention, as these conditioning methods for diffusion models have promising performance in the image generation domain).

Lastly, we found that diffusion-based language models work well with limited data and no external knowledge or pre-trained weights, thus these models may have great potential under low-resource settings, but we didn't apply them to any real low-resource settings (such as low-resource languages, rare domains, etc) in this paper, and we would like to do that in the future to explore the full potential of diffusion-based language models.
\section{Conclusions}

In this paper, we explored the capabilities of diffusion-based models on fine-grained text style transfer, a task that requires a high level of control over generated text, with no external knowledge or pre-trained weights and with very limited training data. Our diffusion-based language model, which builds upon DiffuSeq~\cite{gong2022diffuseq}, achieves state-of-the-art performance on all transfers as well as composition of transfers, outperforming all previous works on this dataset, including ones that uses pre-trained weights, word embeddings, and external grammar parsers. It is even on par with human performance on many transfers. Therefore, our model is a great step towards solving automated fine-grained text style transfer.

Moreover, our work, together with previous works such as Diffusion-LM~\cite{li2022diffusion}, demonstrates that diffusion-based language models could have great potential in controllable text generation under low-resource settings. Under low-resource settings (such as rarely spoken language or uncommon tasks), it would be difficult to find existing large language models or pre-trained weights, and available training data will likely be very limited, so most approaches based on finetuning existing models or large amounts of training will not work well, and diffusion-based language models could be an alternative to consider.

\section*{Acknowledgement}

This work is supported in part by grants from NSF IIS 1453651, NIH K12 NS080223, Cook Family Brain Tumor Research Fund, Mark Trauner Brain Research Fund: Zenkel Family Foundation, Ian’s Friends Foundation, and the Investigators Awards grant program of Precision Health at the University of Michigan. Any opinions, findings, conclusions, or recommendations expressed in this work are those of the author(s) and do not necessarily reflect the views of the NSF, NIH, Cook Family Brain Tumor Research Fund, Mark Trauner Brain Research Fund: Zenkel Family Foundation, Ian’s Friends Foundation, or Precision Health at the University of Michigan. We are grateful to the reviewers for their helpful review and feedback.

\bibliography{main}

\end{document}

%% file: tables/example.tex
\begin{table*}[t]
\fontsize{9}{11}\selectfont
\setlength\tabcolsep{2.0pt}
\centering
\scalebox{0.83}[0.96]{
\begin{tabular}{cllllc}
\toprule
Aspect & Transfers & Original Sentence & \begin{tabular}[l]{@{}l@{}} Additional \\ Information\end{tabular} & Transformed Sentence & \begin{tabular}[l]{@{}l@{}} Number of Pairs \\ in StylePTB dataset\end{tabular}  \\ \hline

\multirow{12}{*}{Syntax} & To Future Tense & \begin{tabular}[p]{@{}l@{}}She travels to Paris every \\ summer to visit her family.\end{tabular}  &  & \begin{tabular}[p]{@{}l@{}}She \textbf{will} travel to Paris next \\ summer to visit her family.\end{tabular}   & 7272  \\

 \cline{2-6}
  
 & To Present Tense & \begin{tabular}[l]{@{}l@{}}She had been studying \\architecture for five years.\end{tabular} &  & \begin{tabular}[l]{@{}l@{}}She \textbf{has} been studying architecture for \\five years.\end{tabular} & 4365  \\
 
\cline{2-6}
 & To Past Tense & \begin{tabular}[c]{@{}l@{}}He walks to the store every day.\end{tabular} &  & \begin{tabular}[c]{@{}l@{}}He \textbf{walked} to the store every day.\end{tabular} & 4422  \\
   
 \cline{2-6}
 & Activate to Passive & \begin{tabular}[c]{@{}l@{}}The cat chased the mouse.\end{tabular} &  & \begin{tabular}[c]{@{}l@{}}The mouse \textbf{was chased} by the cat.\end{tabular} & 2808  \\
   
\cline{2-6}
 & Passive to Activate & \begin{tabular}[c]{@{}l@{}}The proposal was approved by \\ the committee yesterday.\end{tabular} &  & \begin{tabular}[c]{@{}l@{}} The committee \\ \textbf{approved}  the proposal yesterday.\end{tabular} & 2808  \\
   
\cline{2-6}
 & PP Front to Back & \begin{tabular}[l]{@{}l@{}}Having watched the movie, \\ they left the theater.\end{tabular} &  & \begin{tabular}[l]{@{}l@{}}They left the theater after \\ \textbf{having watched the movie}.\end{tabular} & 467  \\
 \cline{2-6}

 & PP Back to Front & \begin{tabular}[l]{@{}l@{}}They have been planning \\ their vacation for months.\end{tabular} &  & \begin{tabular}[l]{@{}l@{}}\textbf{For months}, they have been \\ planning their vacation.\end{tabular} & 467  \\

 \hline
 \multirow{8}{*} {Semantic} & 
 {ADJ/ADV Removal} & \begin{tabular}[l]{@{}l@{}}The \textbf{extremely talented} musician\\ played a \textbf{beautiful} melody on the piano.\end{tabular} &  & \begin{tabular}[l]{@{}l@{}}The musician played \\ a melody on the piano.\end{tabular} & 4863  \\
\cline{2-6}
& PP Removal & \begin{tabular}[l]{@{}l@{}}She had been studying \textbf{for hours} \\ before taking the test.\end{tabular} &  & \begin{tabular}[l]{@{}l@{}}She had been studying \\ before taking the test.\end{tabular} & 4767  \\
\cline{2-6}
& Substatement Removal & \begin{tabular}[l]{@{}l@{}} He was unhappy \textbf{that he had failed}. \\ \textbf{the exam}\end{tabular} & & \begin{tabular}[l]{@{}l@{}} He was unhappy.\end{tabular} & 1345  \\
\cline{2-6}
& Infomation Addition & \begin{tabular}[l]{@{}l@{}}The stock was up three\\ percent according to the man.\end{tabular} & "man", "lazy" & \begin{tabular}[l]{@{}l@{}}The stock was up three percent\\ according to the \textbf{lazy} man.\end{tabular} & 2114  \\

\hline
 \multirow{3}{*} {Thematics} & 
 {Verb/Action Emphasis} &  \begin{tabular}[l]{@{}l@{}}She reads books in pastime. \end{tabular} & "read" & \begin{tabular}[l]{@{}l@{}} \textbf{Reading} books is her favorite pastime. \end{tabular} & 1201  \\
\cline{2-6}
& Adjective Emphasis & \begin{tabular}[l]{@{}l@{}}The scenic forest is \\ Michele's favorite place.\end{tabular} & "scenic" & \begin{tabular}[l]{@{}l@{}} Michele's favorite forest is \textbf{scenic}.\end{tabular} & 696  \\

\hline

\end{tabular}
}
\caption{The 13 non-lexical fine-grained text style transfers from the StylePTB dataset~\cite{lyu2021styleptb}. We present one example sentence pair before/after each transfer, as well as the total number of sentence pairs available for each transfer in StylePTB. As we can see, the transfers require changing one specific stylistic aspect of the sentence while leaving all other aspects unchanged, and the amount of data available for training is limited (compared to the typical amount of data required to train large language models nowadays).
   \vspace{-2mm} }

\label{tab:example}
\end{table*}

%% file: tables/res_short.tex
\begin{table*}[]
\fontsize{9}{11}\selectfont
\setlength\tabcolsep{5.0pt}
\centering
\vspace{-6mm}
\scalebox{0.92}[0.87]{
\begin{tabular}{llccccccc}
\toprule
\textbf{Easy} Transfers & Baseline Model & \multicolumn{1}{l}{BLEU-1} & \multicolumn{1}{l}{BLEU-2} & \multicolumn{1}{l}{BLEU-3} & \multicolumn{1}{l}{BLEU-4} & \multicolumn{1}{l}{METEOR} & \multicolumn{1}{l}{ROUGE\_L} & \multicolumn{1}{l}{CiDER} \\
\hline
\multirow{8}{*}{To Future Tense} & \textsc{GPT2} & 0.895 & 0.852 & 0.813 & 0.778 & 0.540 & 0.899 & 7.709 \\
 & \textsc{Seq2seq} & 0.527 & 0.368 & 0.261 & 0.188 & 0.173 & 0.531 & 1.525 \\
 & \textsc{RetrieveEdit} & 0.899 & 0.854 & 0.815 & 0.778 & 0.531 & 0.901 & 7.731 \\ 
  & \textsc{Steering Vector} & 0.699 & - & - & - & - &  & - \\
  & \textsc{TAILOR} & 0.873 &- &  -& - &-  & - &-  \\
  & \textsc{Diffuseq} & 0.976 & 0.956 & 0.937 & 0.917 & 0.646 & 0.973 & 9.145 \\
  & \textsc{Diffuseq MultiTask} & \textbf{0.985} & \textbf{0.972} & \textbf{0.959} & \textbf{0.946} & \textbf{0.677} & \textbf{0.983} & \textbf{9.454} \\ 
 \cline{2-9}
 & \textsc{Human} & 0.954 & 0.915 & 0.884 & 0.855 & 0.636 & 0.964 & 9.174 \\
 \hline
 \multirow{8}{*}{To Past Tense} & \textsc{GPT2} & 0.836 & 0.776 & 0.722 & 0.674 & 0.484 & 0.842 & 6.700 \\
 & \textsc{Seq2seq} & 0.478 & 0.313 & 0.204 & 0.133 & 0.155 & 0.490 & 1.374 \\
 & \textsc{RetrieveEdit} & 0.935 & 0.903 & 0.873 & 0.847 & 0.606& 0.933 & 8.358 \\ 
  & \textsc{Steering Vector} & 0.478 &- &-  &  -& - & - &- \\
  & \textsc{TAILOR} & 0.711 & -& - & - & - & - & - \\
   & \textsc{Diffuseq} & 0.973 & 0.959 & 0.946 & 0.932 & 0.697 & 0.976 & 9.352 \\
  & \textsc{Diffuseq MultiTask} & \textbf{0.986} & \textbf{0.977} & \textbf{0.968} & \textbf{0.958} & \textbf{0.709} & \textbf{0.987} & \textbf{9.588} \\ 
 \cline{2-9}
 & \textsc{Human} & 0.974 & 0.957 & 0.939 & 0.916 & 0.709 & 0.982 & 9.549 \\ \hline
\multirow{8}{*}{To Present Tense} & \textsc{GPT2} & 0.754 & 0.663 & 0.586 & 0.524 & 0.412 & 0.772 & 5.293 \\
 & \textsc{Seq2seq} & 0.516 & 0.361 & 0.267 & 0.210 & 0.190 & 0.518 & 1.819 \\
 & \textsc{RetrieveEdit} & 0.909 & 0.870 & 0.830 & 0.793 & 0.599& 0.916 & 7.987 \\ 
  & \textsc{Steering Vector} & 0.692 & -& - & - & - & - & - \\
  & \textsc{TAILOR} & 0.884 & -& -& - & - & - & - \\
    & \textsc{Diffuseq} & 0.965 & 0.948 & 0.932 &0.916 & 0.713 & 0.964 & 9.072 \\
  & \textsc{Diffuseq MultiTask} & \textbf{0.975} & \textbf{0.961} & \textbf{0.947} & \textbf{0.933} & \textbf{0.719} & \textbf{0.977} & \textbf{9.310} \\ 
 
 \cline{2-9}
 & \textsc{Human} & 0.969 & 0.952 & 0.936 & 0.918 & 0.745 & 0.979 & 9.501 \\\hline
\multirow{8}{*}{ADJ or ADV Removal} & \textsc{GPT2} & 0.647 & 0.508 & 0.394 & 0.308 & 0.313 & 0.652 & 3.259 \\
 & \textsc{Seq2seq} & 0.450 & 0.274 & 0.172 & 0.112 & 0.140 & 0.469 & 1.171 \\
 & \textsc{RetrieveEdit} & 0.897 & 0.841 & 0.786 & 0.731 & 0.511 & 0.919 & 7.461\\ 
  & \textsc{Steering Vector} & 0.721 & -& - & - &-  & - &-  \\
  & \textsc{TAILOR} & 0.781 &- &  -& - & - &  -&-  \\
  & \textsc{Diffuseq} & 0.903 & 0.809 & 0.731 & 0.664 & 0.488 & 0.888 & 6.708 \\
  & \textsc{Diffuseq MultiTask} & \textbf{0.949} & \textbf{0.908} & \textbf{0.868} & \textbf{0.829} & \textbf{0.563} & \textbf{0.946} & \textbf{8.237} \\ 
 
 \cline{2-9}
 & \textsc{Human} & 0.933 & 0.894 & 0.870 & 0.847 & 0.591 & 0.965 & 8.924 \\
\hline
\end{tabular}
}
\vspace{2mm}
\scalebox{0.92}[0.87]{
\begin{tabular}{llccccccc}
\toprule
\textbf{Medium} Transfers & Baseline Model & \multicolumn{1}{l}{BLEU-1} & \multicolumn{1}{l}{BLEU-2} & \multicolumn{1}{l}{BLEU-3} & \multicolumn{1}{l}{BLEU-4} & \multicolumn{1}{l}{METEOR} & \multicolumn{1}{l}{ROUGE\_L} & \multicolumn{1}{l}{CiDER} \\
 \hline
\multirow{8}{*}
{PP Front to Back} & \textsc{GPT2} & 0.398 & 0.210 & 0.081 & 0.001 & 0.184 & 0.406 & 0.886 \\
 & \textsc{Seq2seq} & 0.393 & 0.280 & 0.207 & 0.161 & 0.162 & 0.391 & 1.492 \\
 & \textsc{RetrieveEdit} & 0.541 & 0.423 & 0.301 & 0.176 & 0.247 & 0.547 & 2.536\\ 
  & \textsc{Steering Vector} & 0.819 & -&-  & - & - &  -& - \\
  & \textsc{TAILOR} & 0.842 &- & - & - &  -& - &-  \\
   & \textsc{Diffuseq} & 0.605 & 0.409 & 0.301 & 0.247 & 0.271 & 0.514 & 2.273 \\
  & \textsc{Diffuseq MultiTask} & \textbf{0.978} & \textbf{0.931} & \textbf{0.893} & \textbf{0.856} & \textbf{0.567} & \textbf{0.901} & \textbf{8.374} \\ 
 
 \cline{2-9}
 & \textsc{Human} & 0.965 & 0.959 & 0.952 & 0.945 & 0.690 & 0.970 & 9.671 
 \\ \hline
 \multirow{6}{*}
{PP Back to Front} & \textsc{GPT2} & 0.407 & 0.241 & 0.091 & 0.001 & 0.166 & 0.406 & 0.931
 \\
 & \textsc{Seq2seq} & 0.298 & 0.157 & 0.090 & 0.060 & 0.112 & 0.284  & 0.606 \\
 & \textsc{RetrieveEdit} & 0.649 & 0.584 & 0.535 & 0.491 & 0.333 & 0.656 & 4.667
 \\ 
   & \textsc{Diffuseq} & 0.603 & 0.400 & 0.291 & 0.242 & 0.266 & 0.514 & 2.255 \\
  & \textsc{Diffuseq MultiTask} & \textbf{0.983} & \textbf{0.944} & \textbf{0.905} & \textbf{0.868} & \textbf{0.610} & \textbf{0.950} & \textbf{8.664} \\ 
 
 \cline{2-9}
 & \textsc{Human} & 1.000 & 1.000 & 1.000 & 1.000 & 1.000 & 1.000 & 10.000
 
 \\ \hline

 \multirow{8}{*}{PP Removal} & \textsc{GPT2} & 0.763 & 0.700 & 0.645 & 0.593 & 0.419 & 0.787 & 6.012 \\
 & \textsc{Seq2seq} & 0.330 & 0.195 & 0.121 & 0.081 & 0.112 & 0.363 & 1.004 \\
 & \textsc{RetrieveEdit} & 0.798 & 0.770 & 0.739 & 0.712 & 0.478& 0.846 & 7.111 \\ 
  & \textsc{Steering Vector} & 0.393 & -&-  &-  & - &-  & - \\
  & \textsc{TAILOR} & 0.717 & -& - &-  & - &  -&  -\\
    & \textsc{Diffuseq} & 0.856 & 0.803 & 0.758 & 0.717 & 0.515 & 0.872 & 7.235 \\
  & \textsc{Diffuseq MultiTask} & \textbf{0.950} & \textbf{0.937} & \textbf{0.919} & \textbf{0.902} & \textbf{0.624} & \textbf{0.948} & \textbf{8.606} \\ 
 
 \cline{2-9}
 & \textsc{Human} & 0.957 & 0.944 & 0.931 & 0.919 & 0.681 & 0.976 & 9.207 \\\hline

\multirow{7}{*}{Substatement Removal} & \textsc{GPT2} & 0.430 & 0.332 & 0.247 & 0.176 & 0.250 & 0.588 & 3.090 \\
 & \textsc{Seq2seq} & 0.317 & 0.192 & 0.110 & 0.001 & 0.100 & 0.368 & 1.041 \\
 & \textsc{RetrieveEdit} & 0.706 & 0.678 & 0.647 & 0.607 & 0.405 & 0.767 & 6.183 \\
  & \textsc{Steering Vector} & 0.120 & -& - & - & - &-  &-  \\
   & \textsc{Diffuseq} & 0.688 & 0.592 & 0.493 & 0.388 & 0.364 & 0.718 & 4.285 \\
  & \textsc{Diffuseq MultiTask} & \textbf{0.884} & \textbf{0.860} & \textbf{0.825} & \textbf{0.781} & \textbf{0.555} & \textbf{0.895} & \textbf{7.165} \\ 
 
 \cline{2-9} 
 & \textsc{Human} & 0.731 & 0.720 & 0.705 & 0.685 & 0.607 & 0.788 & 7.691 \\
 \hline
\multirow{7}{*}{Information Addition} & \textsc{GPT2} & 0.479 & 0.305 & 0.189 & 0.121 & 0.207 & 0.475 & 1.359 \\
 & \textsc{Seq2seq} & 0.345 & 0.180 & 0.094 & 0.053 & 0.098 & 0.335 & 0.632 \\
  & \textsc{Steering Vector} & 0.772 & -& - & - &  -&-  &-  \\
 & \textsc{RetrieveEdit} & 0.493 & 0.396 & 0.328 & 0.275 & 0.284 & 0.603 & 3.401 \\ 
    & \textsc{Diffuseq} & 0.809 & 0.572 & 0.420 & 0.3081
    & 0.3829
    & 0.676 & 3.439
    \\
  & \textsc{Diffuseq MultiTask} & \textbf{0.911} & \textbf{0.800} & \textbf{0.706} & \textbf{0.623} & \textbf{0.483} & \textbf{0.835} & \textbf{6.038} \\ 
 
 \cline{2-9}
 & \textsc{Human} & 0.846 & 0.762 & 0.690 & 0.624 & 0.521 & 0.892 & 6.863 \\
\hline
\end{tabular}
}
\vspace{-2mm}

\caption{Evaluation results on easy and medium transfers. DiffuSeq Multitask achieves State of the art performance in every metric, and is on par with human performance. \vspace{-2mm}}
\label{tab:bas_short}
\end{table*}

%% file: tables/full_hard.tex
\begin{table*}[t]
\fontsize{9}{11}\selectfont
\setlength\tabcolsep{5.0pt}
\centering
\vspace{-6mm}

\scalebox{0.92}[0.92]{
\begin{tabular}{llccccccc}
\toprule
\textbf{Hard} Transfers & Baseline Model & \multicolumn{1}{l}{BLEU-1} & \multicolumn{1}{l}{BLEU-2} & \multicolumn{1}{l}{BLEU-3} & \multicolumn{1}{l}{BLEU-4} & \multicolumn{1}{l}{METEOR} & \multicolumn{1}{l}{ROUGE\_L} & \multicolumn{1}{l}{CiDER} \\ \hline
\multirow{10}{*}{Active To Passive} & \textsc{GPT2} & 0.476 & 0.329 & 0.238 & 0.189 & 0.216 & 0.464 & 1.820 \\
 & \textsc{Seq2seq} & 0.373 & 0.220 & 0.141 & 0.103 & 0.131 & 0.345 & 0.845 \\
 & \textsc{RetrieveEdit} & 0.681 & 0.598 & 0.503 & 0.427 & 0.383 & 0.663 & 4.535 \\ 
  & \textsc{Steering Vector} & 0.666 & -& - &  -&  -&  -& - \\
  & \textsc{TAILOR} & 0.556 & -&  -& - &-  &-  & - \\
  & \textsc{Neural QCFG} & 0.431 & 0.637& 0.548 & 0.472 & 0.415 &  0.695 & 4.294 \\
  & \textsc{Neural QCFG + copy} & 0.836 & 0.771& 0.713 & 0.662 & 0.499 & 0.803 & 6.410 \\   & \textsc{Diffuseq} & 0.839 & 0.580 & 0.302 & 0.196 & 0.225 & 0.512 & 2.344 \\
  & \textsc{Diffuseq MultiTask} & \textbf{0.918} & \textbf{0.835} & \textbf{0.752} & \textbf{0.681} & \textbf{0.521} & \textbf{0.844} & \textbf{6.913} \\

 \cline{2-9}
 & \textsc{Human} &  0.931 & 0.881 & 0.835 & 0.795 & 0.587 & 0.905 & 8.603 \\
 \hline

\multirow{7}{*}{Passive To Active} & \textsc{GPT2} & 0.433 & 0.271 & 0.167 & 0.120 & 0.191 & 0.434 & 1.329 \\
 & \textsc{Seq2seq} & 0.339 & 0.214 & 0.160 & 0.132 & 0.126 & 0.331 & 1.062 \\
 & \textsc{RetrieveEdit} & 0.714 & 0.659 & 0.559 & 0.474 & 0.397 & 0.732 & 5.024 \\ 
  & \textsc{Steering Vector} & 0.574  & -& - &  -&  -&  -& - \\   & \textsc{Diffuseq} & 0.829 & 0.550 & 0.282 & 0.192 & 0.205 & 0.502 & 2.224\\
  & \textsc{Diffuseq MultiTask} & \textbf{0.955} & \textbf{0.896} & \textbf{0.834} & \textbf{0.777} & \textbf{0.555} & \textbf{0.913} & \textbf{8.028} \\

 \cline{2-9}
 & \textsc{Human} &  0.977 & 0.962 & 0.942 & 0.919 & 0.685 & 0.973 & 9.409 \\
 \hline
 
\multirow{9}{*}{Adjective Emphasis} & \textsc{GPT2} & 0.263 & 0.079 & 0.028 & 0.000 & 0.112 & 0.188 & 0.386 \\
 & \textsc{Seq2seq} & 0.187 & 0.058 & 0.018 & 0.000 & 0.059 & 0.179 & 0.141 \\
 & \textsc{RetrieveEdit} & 0.387 & 0.276 & 0.211& 0.164 & 0.193& 0.369 & 1.679 \\ 
  & \textsc{Steering Vector} & 0.774 & -&-  &-  &-  & - &  -\\
  & \textsc{Neural QCFG} & 0.348 & 0.178& 0.062 & 0.000 & 0.162 &  0.317 & 0.667 \\
  & \textsc{Neural QCFG + copy} & 0.676 & 0.506& 0.393 & 0.316 & 0.373 & 0.683 & 3.424 \\
 & \textsc{DiffuSeq} & 0.620 & 0.382 & 0.215& 0.152 & 0.243& 0.335 & 2.231\\ 
   & \textsc{Diffuseq MultiTask} & \textbf{0.775} & \textbf{0.600} & \textbf{0.477} & \textbf{0.386} & \textbf{0.423} & \textbf{0.673} & \textbf{4.007} \\

 \cline{2-9}
 & \textsc{Human} & 0.834 & 0.753 & 0.679 & 0.611 & 0.522 & 0.811 & 6.796 \\
 \hline
\multirow{9}{*}{Verb/Action Emphasis} & \textsc{GPT2} & 0.309 & 0.170 & 0.095 & 0.041 & 0.140 & 0.292 & 0.593 \\
 & \textsc{Seq2seq} & 0.289 & 0.127 & 0.066 & 0.038 & 0.098 & 0.275 & 0.300 \\
 & \textsc{RetrieveEdit} & 0.416 & 0.284 & 0.209 & 0.148 & 0.223 & 0.423 & 1.778\\  
  & \textsc{Steering Vector} & 0.548 & -&  -& - &-  & - &-  \\
  & \textsc{Neural QCFG} & 0.431 & 0.250& 0.14 & 0.073 & 0.219 &  0.408 & 1.097 \\
  & \textsc{Neural QCFG + copy} & 0.664 & 0.512 & \textbf{0.407} & \textbf{0.319} & 0.370 & 0.589 & \textbf{3.227} \\
 & \textsc{DiffuSeq} & 0.453 & 0.210 & 0.101 & 0.054 & 0.205 & 0.379 & 0.785\\
   & \textsc{Diffuseq MultiTask} & \textbf{0.693} & \textbf{0.516} & 0.370 & 0.261 & \textbf{0.373} & \textbf{0.596} & 2.950 \\

 \cline{2-9}
 & \textsc{Human} & 0.649 & 0.569 & 0.493 & 0.421 & 0.433 & 0.693 & 5.668 \\
\hline
\end{tabular}
}\vspace{-1mm}
\caption{Evaluation results on hard transfers. Diffuseq Multitask achieves State-of-the-art performance on most metrics, and is only slightly behind Neural QCFG + copy on some metrics. \vspace{-2mm}}
\label{tab:full_shard}
\end{table*}

%% file: tables/comp_full.tex
\begin{table*}[t]
\fontsize{9}{11}\selectfont
\setlength\tabcolsep{2.0pt}
\centering
\scalebox{0.97}{
\begin{tabular}{lllccccccc}
\toprule
Dataset & Transfers & Model & BLEU-1 & BLEU-2 & BLEU-3 & BLEU-4 & METEOR & ROUGE\_L & CiDER \\ \hline
\multirow{18}{*}{\begin{tabular}[c]{@{}l@{}}Tense\\ +\\ Voice\end{tabular}} & \multirow{3}{*}{\begin{tabular}[c]{@{}l@{}}ToPast+\\ ActiveToPassive\end{tabular}} & \textsc{SeqGPT} & 0.332 & 0.155 & 0.057 & 0.024 & 0.144 & 0.300 & 0.636 \\
 & & CS-GPT & 0.409 & 0.238 & 0.133 & 0.064 & 0.180 & 0.378 & 1.029 \\
  & & \textsc{DiffuSeq} & \textbf{0.744} & \textbf{0.555} & \textbf{0.420} & \textbf{0.324} & \textbf{0.353} & \textbf{0.656} & \textbf{3.753} \\
 
 \cline{2-10}
 & \multirow{3}{*}{\begin{tabular}[c]{@{}l@{}}ToFuture+\\ ActiveToPassive\end{tabular}} & \textsc{SeqGPT} & 0.391 & 0.222 & 0.120 & 0.065 & 0.167 & 0.373 & 0.866 \\
 & & CS-GPT & 0.496 & 0.340 & 0.240 & 0.185 & 0.217 & 0.479 & 1.800 \\
   & & \textsc{DiffuSeq} & \textbf{0.821} & \textbf{0.705} & \textbf{0.615} & \textbf{0.542} & \textbf{0.414} & \textbf{0.762} & \textbf{5.281} \\
 
 \cline{2-10}
 & \multirow{3}{*}{\begin{tabular}[c]{@{}l@{}}ToFuture+\\ PassiveToActive\end{tabular}} & \textsc{SeqGPT} & 0.401 & 0.212 & 0.097 & 0.048 & 0.163 & 0.385 & 0.888 \\
 & & CS-GPT & 0.528 & 0.364 & 0.259 & 0.197 & 0.234 & 0.524 & 2.020 \\
   & & \textsc{DiffuSeq} & \textbf{0.744} & \textbf{0.555} & \textbf{0.420} & \textbf{0.324} & \textbf{0.353} & \textbf{0.656} & \textbf{3.753} \\
   
 \cline{2-10}
 & \multirow{3}{*}{\begin{tabular}[c]{@{}l@{}}ToPast+\\ PassiveToActive\end{tabular}} & \textsc{SeqGPT} & 0.381 & 0.210 & 0.098 & 0.045 & 0.156 & 0.368 & 0.876 \\
 & & CS-GPT & 0.474 & 0.297 & 0.175 & 0.099 & 0.206 & 0.473 & 1.513 \\
   & & \textsc{DiffuSeq} & \textbf{0.864} & \textbf{0.772} & \textbf{0.697} & \textbf{0.635} & \textbf{0.460} & \textbf{0.825} & \textbf{6.519} \\
   
 \cline{2-10} 
 & \multirow{3}{*}{\begin{tabular}[c]{@{}l@{}}ToPresent+\\ PassiveToActive\end{tabular}} & \textsc{SeqGPT} & 0.348 & 0.189 & 0.085 & 0.037 & 0.142 & 0.343 & 0.745 \\
 & & \textsc{CS-GPT} & 0.523 & 0.366 & 0.264 & 0.210 & 0.243 & 0.522 & 2.118 \\
   & & \textsc{DiffuSeq} & \textbf{0.797} & \textbf{0.686} & \textbf{0.603} & \textbf{0.536} & \textbf{0.414} & \textbf{0.756} & \textbf{5.378} \\
   
 \cline{2-10} 
 & \multirow{3}{*}{\begin{tabular}[c]{@{}l@{}}ToPresent+\\ ActiveToPassive\end{tabular}} & \textsc{SeqGPT} & 0.396 & 0.256 & 0.177 & 0.136 & 0.179 & 0.384 & 1.209 \\
 & & \textsc{CS-GPT} & 0.503 & 0.358 & 0.271 & 0.223 & 0.233 & 0.491 & 2.118 \\
   & & \textsc{DiffuSeq} & \textbf{0.878} & \textbf{0.787} & \textbf{0.715} & \textbf{0.656} & \textbf{0.482} & \textbf{0.849} & \textbf{6.823} \\

 \hline
 \multirow{9}{*}{\begin{tabular}[c]{@{}l@{}}Tense\\ +\\ PP \\ Removal\end{tabular}} & \multirow{3}{*}{\begin{tabular}[c]{@{}l@{}}ToFuture+\\ PPRemoval\end{tabular}} & \textsc{SeqGPT} & 0.722 & 0.644 & 0.581 & 0.524 & 0.385 & \textbf{0.755} & 5.562 \\
 & & CS-GPT & 0.738  & 0.652 &  0.578 &  0.518  & 0.393 &  0.755  & 5.289 \\
  & & DiffuSeq & \textbf{0.913} &\textbf{0.876} &\textbf{0.841} &\textbf{0.808} & \textbf{0.557} & \textbf{0.911} & \textbf{7.906} \\
 \cline{2-10} 
 & \multirow{3}{*}{\begin{tabular}[c]{@{}l@{}}ToPast+\\ PPRemoval\end{tabular}} & \textsc{SeqGPT} & 0.714 & 0.640 & 0.573 & 0.510 & 0.374 & 0.724 & 5.152 \\
 & & CS-GPT & 0.772 &  0.695 &  0.624 &  0.564 &  0.421 &  0.775 &  5.585 \\
  & & DiffuSeq &  \textbf{0.911} & \textbf{0.881} & \textbf{0.849} & \textbf{0.818} & \textbf{0.568} & \textbf{0.908} & \textbf{7.825}\\
 \cline{2-10} 
 & \multirow{3}{*}{\begin{tabular}[c]{@{}l@{}}ToPresent+\\ PPRemoval\end{tabular}} & \textsc{SeqGPT} & 0.618 & 0.518 & 0.435 & 0.368 & 0.338 & 0.663 & 4.119 \\
 
 & & CS-GPT & 0.709 & 0.609 & 0.523 & 0.446 & 0.718 & 0.718 & 4.588 \\
 & & DiffuSeq & \textbf{0.908} & \textbf{0.859} & \textbf{0.820} & \textbf{0.788} & \textbf{0.558} & \textbf{0.895} & \textbf{7.439}\\
\hline

\end{tabular}
}
\vspace{-1mm}
\caption{Results on compositions of Tense + Voice transfers and Tense + PP Removal Transfers. DiffuSeq was able to outperform all 3 baselines from~\cite{lyu2021styleptb} by a large margin. }
\vspace{-2mm}
\label{tab:comp_full}
\end{table*}